\documentclass{article}

\usepackage{microtype}
\usepackage{graphicx}
\usepackage{subfigure}
\usepackage{booktabs} 
\usepackage{graphicx}
\usepackage{multirow}
\usepackage{pifont}
\usepackage[table,xcdraw]{xcolor}
\usepackage{textcomp}
\usepackage{hyperref}
\usepackage{tcolorbox}

\hypersetup{
    colorlinks=true,       
    linkcolor=cyan,        
    citecolor=red,         
    filecolor=magenta,     
    urlcolor=cyan          
}

\definecolor{lightblue}{rgb}{0.18,0.45,0.71} 

\usepackage[accepted]{icml2025}
\usepackage{amsmath}
\usepackage{amssymb}
\usepackage{mathtools}
\usepackage{amsthm}

\usepackage[capitalize,noabbrev]{cleveref}

\theoremstyle{plain}

\theoremstyle{definition}

\theoremstyle{remark}

\usepackage[textsize=tiny]{todonotes}

\begin{document}

\twocolumn[
\icmltitle{Slow-Fast Architecture for Video Multi-Modal Large Language Models}



\icmlsetsymbol{equal}{$\dagger$}
\icmlsetsymbol{intern}{*}
\icmlsetsymbol{first}{1}
\icmlsetsymbol{second}{2}
\icmlsetsymbol{third}{3}
\icmlsetsymbol{fourth}{4}

\begin{icmlauthorlist}
\icmlauthor{Min Shi}{first,intern}
\icmlauthor{Shihao Wang}{third,intern}
\icmlauthor{Chieh-Yun Chen}{first}
\icmlauthor{Jitesh Jain}{first}
\icmlauthor{Kai Wang}{first}\\
\icmlauthor{Jinjun Xiong}{fourth} 
\icmlauthor{Guilin Liu}{second}
\icmlauthor{Zhiding Yu}{equal,second}
\icmlauthor{Humphrey Shi}{equal,first}
\end{icmlauthorlist}

\centering{{\small $^{\mathrm{1}}$SHI Labs @ Georgia Tech \quad $^{\mathrm{2}}$NVIDIA \quad $^{\mathrm{3}}$HKPU \quad $^{\mathrm{4}}$SUNY Buffalo}} \\
\centering{{\small \textbf{\texttt{\href{https://github.com/SHI-Labs/Slow-Fast-Video-Multimodal-LLM}{github.com/SHI-Labs/Slow-Fast-Video-Multimodal-LLM}}}}}
\icmlcorrespondingauthor{Zhiding Yu}{zhidingy@nvidia.com}
\icmlcorrespondingauthor{Humphrey Shi}{shi@gatech.edu}

\quad
\icmlkeywords{Multi-modal Large Language Model, Video Understanding}

\vskip 0.3in
]



\printAffiliationsAndNotice{* Work partially done during an internship at NVIDIA. $\dagger$ Equal advising. Correspondence to: Humphrey Shi \small{\textless shi@gatech.edu\textgreater}, Zhiding Yu \small{\textless zhidingy@nvidia.com\textgreater}}  

\begin{abstract}
Balancing temporal resolution and spatial detail under limited compute budget remains a key challenge for video-based multi-modal large language models (MLLMs). 
Existing methods typically compress video representations using predefined rules before feeding them into the LLM, resulting in irreversible information loss and often ignoring input instructions.
To address this, we propose a novel slow-fast architecture that naturally circumvents this trade-off, enabling the use of more input frames while preserving spatial details.
Inspired by how humans first skim a video before focusing on relevant parts, our slow-fast design employs a dual-token strategy:
1) “fast” visual tokens — a compact set of compressed video features — are fed into the LLM alongside text embeddings to provide a quick overview;
2) ``slow'' visual tokens — uncompressed video features — are cross-attended by text embeddings through specially designed hybrid decoder layers, enabling instruction-aware extraction of relevant visual details with linear complexity.
We conduct systematic exploration to optimize both the overall architecture and key components. Experiments show that our model significantly outperforms self-attention-only baselines, extending the input capacity from 16 to 128 frames with just a 3\% increase in computation, and achieving a 16\% average performance improvement across five video understanding benchmarks.
Our 7B model achieves state-of-the-art performance among models of similar size.
Furthermore, our slow-fast architecture is a plug-and-play design that can be integrated into other video MLLMs to improve efficiency and scalability.
\end{abstract}

\vspace{-10pt}
\section{Introduction}
\label{sec:intro}

\begin{figure}[!t]
  \centering
  \includegraphics[width=1.0\linewidth]{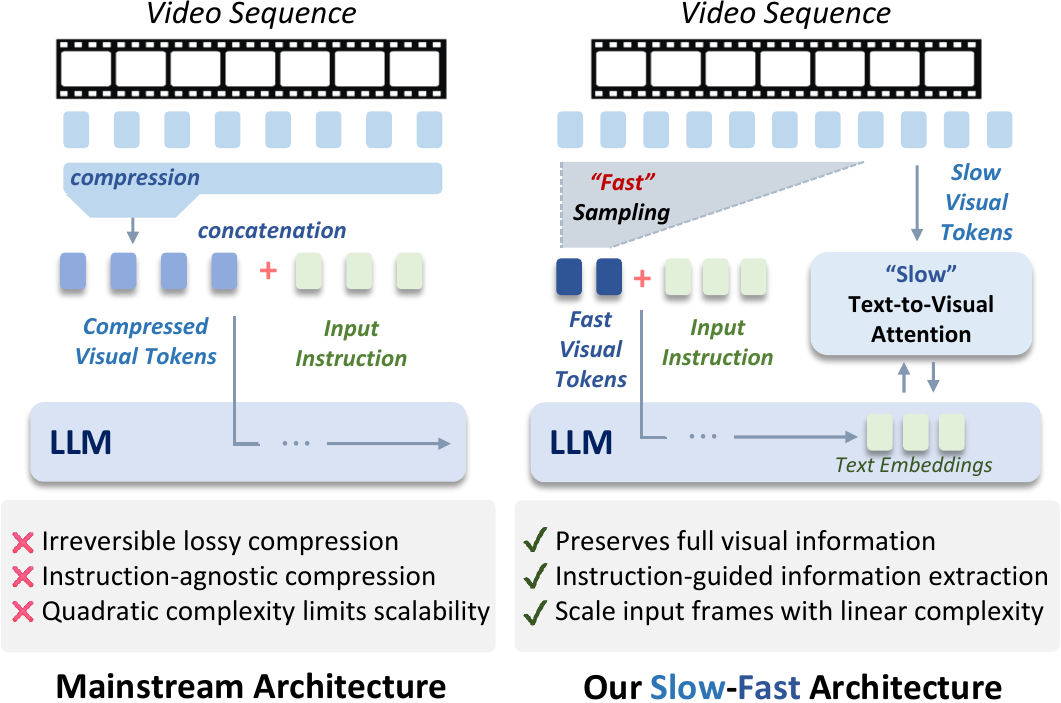}
  \vspace{-5pt}
   \caption{\textbf{Comparison between the mainstream video MLLM architecture and the proposed slow-fast architecture.} 
   Rather than relying on carefully-designed video representation compression strategies, the slow-fast architecture utilizes highly compressed ``fast'' visual tokens as a preview for the LLM while allowing text embeddings to extract relevant information from uncompressed ``slow'' visual tokens via cross-attention. This approach extends a 16-frame baseline to a 96-frame input with only a 2\% increase in computation, yielding a 14\% average performance improvement across five benchmarks.
   }
   \label{fig:paradigm-comparison}
   \vspace{-5pt}
\end{figure}

Video Multi-modal Large Language Models (MLLMs) enable Large Language Models (LLMs) to perform complex reasoning based on encoded video features. Mainstream approaches~\cite{videollama,qwen2vl,chen2024internvl} treat video as a sequence of images, concatenating frame features to construct the video representation fed into the LLM. Due to the LLM’s context length constraints and the redundancy in video data, compressing vision tokens before feeding them into LLMs has become a standard post-processing step~\cite{qwen2vl,liu2024oryx,longvu}. A key challenge in video representation compression is balancing temporal resolution and spatial detail within the limited context length—whether to include more frames or allocate more tokens per frame.

To achieve a better trade-off while preserving spatially and temporally important information, video MLLMs employ various compression strategies, such as feature merging based on similarity~\cite{longvu} or  attention score~\cite{auroracap}, heuristic pooling rules~\cite{slowfastllava,qwen2vl}, and learnable modules~\cite{liu2024oryx,videollama}. 
However, these methods are irreversible once tokens are fed into the LLM and largely agnostic to input instructions, potentially discarding critical visual details relevant to the given task.
An alternative approach reduces complexity by allowing text embeddings to cross-attend to visual features instead of directly feeding visual tokens into the LLM~\cite{flamingo,otter,infimmhd}. However, in this setup, text embeddings interact with visual features only a few times, lacking persistent visual tokens as contextual anchors within the LLM.
Recent studies~\cite{nvlm} indicate that relying solely on cross-attention underperforms compared to direct concatenation when using the same base model and training data.

The challenges outlined above raise a key question: Can we design a paradigm that preserves sufficient visual details while efficiently and effectively integrating them into the LLM?
Inspired by how humans answer video-based questions~\cite{feichtenhofer2019slowfastnetworksvideorecognition}—first gaining a quick overview and then focusing on relevant details—we propose a slow-fast architecture for video MLLMs to achieve this goal, as illustrated in Fig.~\ref{fig:paradigm-comparison}. 
Video features are first compressed into a fixed number of "fast" visual tokens, which are concatenated with text embeddings to provide a quick preview for the LLM. 
Simultaneously, uncompressed ``slow'' visual tokens interact with text embeddings via cross-attention in modified decoder layers, termed \textit{hybrid decoder layers}, at specific positions. 
This design enables the integration of instruction-relevant visual information while maintaining linear complexity with respect to video length, resulting in an efficient and accurate framework.

We refine the design of the hybrid decoder through a series of explorations. Initially, we observe that simply combining existing cross-attention architectures~\cite{flamingo,grattafiori2024llama3herdmodels} with the LLaVA-style~\cite{llava15} framework yields only marginal improvements while remaining computationally intensive.
To address this, we start by streamlining the integration of the cross-attention module. We compare different designs from recent works~\cite{mplugowl3,cogagent} and remove the computation-intensive components.
The results lead to a minimal yet effective design, revealing that placing a cross-attention operation between the self-attention and feed-forward network in the original decoder layer outperforms the conventional approach of incorporating cross-attention as a stand-alone decoder layer~\cite{flamingo,grattafiori2024llama3herdmodels}. 
Secondly, inspired by the findings that text tokens exhibit varying attention weights to visual tokens~\cite{focusllava}, reflecting different demands for visual content, we propose a dynamic gating mechanism to determine how much visual information can be merged into the text embeddings via cross-attention, conditioned on each text token.
Finally, we investigate crucial implementation details, including weight initialization for the cross-attention module, compression methods for fast visual tokens, and scalability to input frames of up to 96 in a single forward pass.

Based on explorations above, we construct a slow-fast architecture which significantly outperforms mainstream LLaVA-style baselines with identical training data. 
With less than 2\% additional computational overhead from cross-attention, our model extends 16-frame baselines to perceive 96 frames, improving the average performance across five video benchmarks by $14\%$.
Despite this efficiency, our 7B model achieves competitive results compared with other top-performing 7B open-source models and sets state-of-the-art performance on several video understanding benchmarks, \textit{e.g.}, $68.5\%$ on TempCompass~\cite{tempcompass}, $67.9\%$ on MLVU~\cite{mlvu}, $57.5\%$ on LongVideoBench~\cite{longvideobench}, and $70.3\%$ on Perception Test~\cite{perceptiontest}.

Our contributions are summarized as follows: \\
$\bullet$ We are the first to propose a slow-fast architecture for video MLLM that enables instruction-aware visual information extraction from uncompressed video representations with linear complexity, allowing our method to scale efficiently while improving performance.\\
$\bullet$ We conduct an in-depth explorations that reveals the crucial design choices and significantly boost performance by an average of 14\% in five video benchmarks.

\begin{figure*}[!t]
   \centering
   \includegraphics[width=1.0\linewidth]{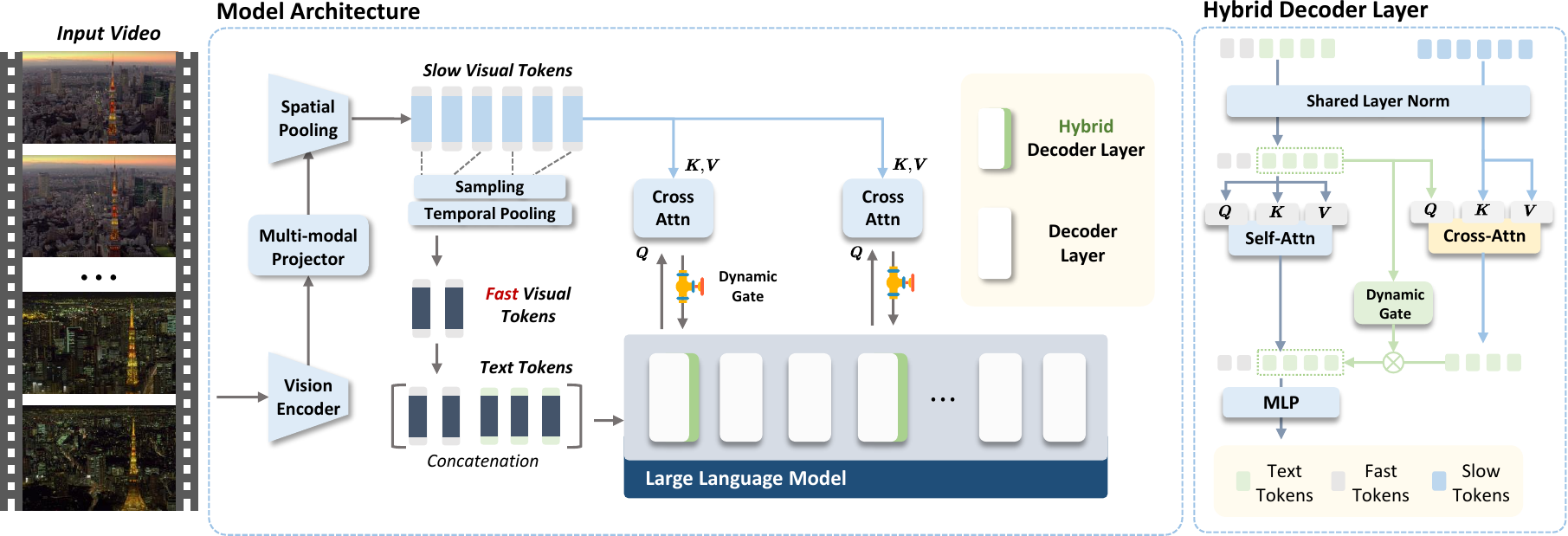}
   \vspace{-0.3cm}
   \caption{
   \textbf{Illustration of the Slow-Fast Architecture and Hybrid Decoder.}
   The video input is first processed into \textit{slow visual tokens} through a vision encoder and projector. 
   These \textit{slow visual tokens} are then condensed into a smaller set of \textit{fast visual tokens} via strided sampling and temporal pooling.
   The \textit{fast visual tokens} are concatenated with text embeddings and fed into the LLM, serving as a preview context.
   Meanwhile, the \textit{slow visual tokens} interact with text embeddings through cross-attention in hybrid decoder layers distributed within the LLM, enabling instruction-aware visual information extraction with linear complexity.
   }
   \label{fig:pipeline-illustration}  
\end{figure*}

\section{Related Work}
\noindent \textbf{Architecture design for MLLM.} 
Multi-modal Large Language Models (MLLMs) enable the LLMs to comprehend more modalities, like images~\cite{llava}, videos~\cite{videochat,videollama}, audio~\cite{qwen2audio}, and 3D data~\cite{3dllm}. 
Two primary approaches have been adopted to integrate multi-modal information from modality-specific encoders. 
The first widely-used approach directly concatenates multi-modal embeddings with text embeddings~\cite{llava,llava15}, which is also followed by a series of top-performing open-source MLLMs~\cite{li2024llavaonevision,chen2024internvl,qwen2vl}.
Under this paradigm, subsequent works further enhancing encoder designs~\cite{chen2024internvl,liu2024oryx}, improving data mixtures and training strategies~\cite{cambrian}, or optimizing training recipes~\cite{prismaticvlm}. 
Alternatively, a second approach use cross-attention between text embeddings and visual tokens within the LLMs to inject visual clues~\cite{flamingo,otter,mplugowl3,infimmhd,grattafiori2024llama3herdmodels} with a linear complexity.
However, under the same training data and strategies, models relying solely on cross-attention demonstrate inferior performance~\cite{nvlm} compared to the first direct concatenation approach.
A few works also try to combine these two paradigms to achieve high-resolution image encoding~\cite{cogagent,nvlm} by feeding the low-resolution snapshots to the LLM while using cross-attention to inject high-resolution image features.
These works share similar implementation and intuition to our slow-fast architecture. Our work further explores the design space of cross-attention integration and demonstrates its potential in video understanding.
 
\noindent \textbf{Video MLLM.}
To encode a video sequence, existing video MLLMs~\cite{videochat,videochatgpt,videollama,qwen2vl,videollama2} typically append features extracted from each frame.
These leads to excessively long sequence length since video can have hundreds of frame, which are undesirable for LLMs especially when video data have a lot of redundancy.
To address this issue, most methods adopt a series of visual representation compression strategies based on pre-defined rules, \textit{e.g.}, heuristic pooling rules~\cite{pllava,zhang2024videoinstructiontuningsynthetic,slowfastllava}, merging similar frames~\cite{auroracap,longvu}, or using learnable modules such as Q-Former~\cite{videochat,videollama} or cross-attention layers~\cite{liu2024oryx} to condense visual representation into fewer tokens.
Some works focus on improving the underlying platform by optimizing attention calculation and distributed training~\cite{longvila}, or extending the context length of LLMs~\cite{longva}. Additionally, memory-based methods~\cite{moivechat} and sliding window techniques~\cite{timechat} have been utilized to consolidate and aggregate information across long video durations.
From a data perspective, several studies~\cite{sharegpt4video,zhang2024videoinstructiontuningsynthetic,llavahound,kangaroo} have enhanced the quality and diversity of video instruction-tuning datasets.
In this paper, we propose a slow-fast architecture which are orthogonal to these efforts and can benefits from each other.

\section{Slow-Fast Architecture for Video MLLM}
As illustrated in Fig.~\ref{fig:pipeline-illustration}, the proposed slow-fast architecture integrates a vision encoder with a multi-modal projector and a large language model, incorporating hybrid layers with gate-controlled cross-attention between text embeddings and visual content. 
Firstly, the vision encoder and multi-modal projector generate a sequence of video features aligned with the text embeddings from the video frames. Unlike common practices~\cite{qwen2vl,li2024llavaonevision,zhang2024videoinstructiontuningsynthetic}, which directly concatenate the compressed visual tokens and text embeddings as input for the LLM, the visual embedding sequence is first compressed into a small set of \textit{fast tokens} via direct sampling and temporal pooling sequentially. 
These \textit{fast tokens} act as a quick preview and are fed into the LLM to establish basic context. Meanwhile, the original, uncompressed visual features are retained as \textit{slow tokens}.
Within the LLM's forward pass, the hybrid layers, which integrate cross-attention upon the pre-trained decoder layer, enable the text embeddings to selectively extract detailed information from the \textit{slow tokens}. A dynamic gate further refines this process by modulating how much information each text token absorbs, dynamically generating gating values from each token. 

\subsection{Encoding Slow and Fast Visual Tokens}
Given a video sequence with $N$ frames, we first process the raw RGB frames using the vision encoder and the multi-modal projector. The resulting frame features undergo a $2 \times 2$ spatial average pooling operation to reduce the token count per frame by 4$\times$. The processed feature sequence is considered as the \textit{slow visual tokens}. Next, we compress the \textit{slow visual tokens} to obtain the \textit{fast visual tokens} by direct sampling and temporal pooling sequentially. Specifically, we sampled the feature sequence every $s$ frames and then applied temporal average pooling with a stride of $t$, resulting in the \textit{fast tokens} $V_f \in \mathbb{R}^{k \times d}$, where $k$ is the number of fast tokens.

\subsection{Hybrid Decoder Layer}
\label{sec:hybrid-decoder-layer}
The core component in our slow-fast architecture is the hybrid decoder layer, which enables text embeddings to interact with uncompressed video features midway through the LLM's forward pass. As shown in the right part of Fig.~\ref{fig:pipeline-illustration}, the hybrid decoder layer extends a pre-trained decoder layer. To seamlessly integrate cross-attention into the decoder, we treat it as analogous to self-attention: self-attention aggregates context from previous tokens, while cross-attention aggregates relevant visual context from the \textit{slow visual tokens}.
Cross-attention also runs in parallel with self-attention, with its outputs merged back into the text embedding via a skip connection. A dynamic gate is added to the skip connection to modulate the output before merging, reducing disturbance and improving training stability.
In the following, we detail the computation of cross-attention and the design of the dynamic gate.

\noindent \textbf{Cross-attention.} 
The input to the hybrid decoder layer consists of hidden states and \textit{slow visual tokens}. Inspired by mPLUG-OWL3~\cite{mplugowl3}, both are first processed through a shared layer normalization layer.
After the layer normalization, there are two attention mechanisms in parallel. All hidden states participate in the self-attention operation. For cross-attention, however, the query is restricted to textual tokens, excluding the \textit{fast visual tokens}. Hence, the queries corresponding to the textual tokens are extracted from the self-attention layer (re-use the query for cross-attention).
Denote this query as $Q_t \in \mathbb{R}^{n \times d}$, where $n$ represents the number of textual tokens, and $d$ is the hidden dimension (head dimensions are omitted for simplicity). Let the normalized \textit{slow visual features} be $V_s\in \mathbb{R}^{m \times d}$, where $m$ denotes the number of slow visual tokens. Then the cross-attention output $X'$ is calculated as:
\begin{equation}
\label{eq:cross-attention-calculation}
    X' = \mathrm{MHCA}(Q_t, W_kV_s, W_vV_s),
\end{equation}
where $\mathrm{MHCA}(q,k,v)$ represents the multi-head cross-attention~ \cite{attentionisallyouneed}, with $q$, $k$, and $v$ serving as the query, key, and value, respectively. $W_k$ and $W_v$ are two learnable projection matrices for the cross-attention module. In practice, the hidden dimension and number of attention heads in the cross-attention layer are kept consistent with the configuration of the pre-trained self-attention layer. 
We also experiment with allowing all the hidden states, including the \textit{fast visual tokens}, to attend to the \textit{slow visual tokens}, as discussed in Sec.~\ref{sec:exp-ablation}. However, our findings indicate that limiting the cross-attention to only the textual tokens results in better performance and efficiency.

\noindent \textbf{Dynamic gate with warm-up.} 
Although the output from the cross-attention layer, $X'$, aggregates the visual context for each text token, it can also introduce disturbances into the pre-trained LLM. A common approach to address this issue is to use a learnable scalar as the gate~\cite{flamingo,grattafiori2024llama3herdmodels} that modulates the extent to which the cross-attention output is merged back into the text embeddings. This static gate assigns an identical weight to all tokens, regardless of their context or importance. However, the necessity of attending to visual contexts can vary significantly across tokens and input instructions, as indicated by varying attention scores~\cite{focusllava}. To address this limitation, we propose a \textit{dynamic gate with warm-up} mechanism. This dynamic gate design allows the model to determine how much of the cross-attention output should be incorporated into the text embeddings for each text token. The warm-up mechanism further stabilizes training at the initial stage.

Specifically, a single linear layer followed by a \texttt{tanh} activation is applied to the text embeddings. This generates a tensor $g_d \in \mathbb{R}^{n}$, which assigns a dynamic gate value to each text token. To mitigate the impact of the cross-attention branch during the initial stages of training, a static learnable warm-up factor $g_s$, initialized to zero, is also introduced. With these components, the cross-attention output $X'$ is merged into the text embedding $X_t$ as:
\begin{equation}
    \label{eq:cross-attention-gating}
    X_t = X_t + X'\circ g_d \cdot g_s\,,
\end{equation}
where $\circ$ denotes the element-wise product. 
Cross-attention updates only $X_t$, the pure text component of the hidden states, while the \textit{fast visual tokens} remain unchanged during this process.
Note that the hidden state is also updated with context from the self-attention output like in the standard decoder layer, where the \textit{fast visual tokens} are included.

\noindent \textbf{Weight initialization.}
For the hybrid decoder layer, the original parameters are retained from the pre-trained models. The gate-related parameters, \textit{i.e.}, the gate linear projection layer is randomly initialized. The key-value projection matrices $W_k$ and $W_v$ of the cross-attention operation are initialized using the weights from the corresponding self-attention layer. In practice, we find this can facilitate training and ensure smoother convergence.

\subsection{Implementation Details}

\noindent \textbf{Model architecture.} 
We adopt ConvNeXt-XXL~\cite{convnext} pre-trained by OpenCLIP~\cite{openclip} as the vision encoder, which downsamples input images by $32$ times. The input resolution for both image and video inputs is set to $576$, resulting in $324$ tokens for each image. For image inputs, the feature map is kept at its original resolution, and the image feature maps serve as both the \textit{fast} and \textit{slow} tokens within the hybrid architecture.
For video inputs, the feature map of each frame is further compressed via average pooling, reducing the token count to $81$ per frame. To account for training samples with low frame counts, we enforce a minimum of $16$ input fast frames during the compression process. 
For the language model, we use Qwen2-7B~\cite{qwen2technicalreport}, a pre-trained large language model with $28$ transformer decoder layers. 
Four hybrid decoder layers are distributed within the LLM's decoder layers at indices $[0, 8, 16, 24]$.

\noindent \textbf{Training recipe.} 
Following common practices~\cite{llava}, we employ a two-stage training strategy for our model. In the first stage, the model is trained on image and video caption datasets. During this stage, only the multi-modal projector and the cross-attention-related parameters, \textit{i.e.}, the key and value projection matrices and the dynamic gate, are updated. The training batch size is set to $256$. The learning rate for the multi-modal projector is set to $1 \times 10^{-3}$, while the learning rate for the cross-attention-related modules is $2 \times 10^{-4}$. Empirically, we find that a higher learning rate for the cross-attention modules leads to training instability. In the second stage, all parameters of the model are fine-tuned using multi-modal conversation datasets across diverse tasks and data formats. This stage incorporates a mixture of text, image, and video data. Here, the batch size is set to $128$, and the learning rate is reduced to $2 \times 10^{-5}$.

\noindent \textbf{Training data.} We construct the training data mixture using open-source datasets. In the first pre-training stage, we combine 537k video caption samples from LLaVA-Video-178k~\cite{zhang2024videoinstructiontuningsynthetic} with 558k image captions from LLaVA-1.5~\cite{llava15}.
In the second stage, we mix pure text, image, and video instruction tuning datasets. Table~\ref{tab:data-mixture-source} provides a detailed breakdown of the data sources and sample sizes. The primary components include 1.4M video data from LLaVA-Video-178k~\cite{zhang2024videoinstructiontuningsynthetic} and 1.2M image data from LLaVA-OneVision~\cite{li2024llavaonevision}. 
Following VideoChat2~\cite{videochat2mvbench}, we also incorporate their conversation data alongside with TGIF~\cite{TGIF}, SthSthv2~\cite{sthsth}, Kinetics-710~\cite{kinetics710}, and Ego4D~\cite{ego4d} into the training mixture.
For the second stage of all ablation studies, due to resource limitations, we use a reduced dataset comprising 665k instruction tuning samples from LLaVA-1.5~\cite{llava15} and 1.4M video instruction tuning samples from LLaVA-Video-178k.

\section{Experiments}
In this section, we present a comparison with recent state-of-the-art video MLLMs, followed by controlled ablation studies and qualitative examples. 
We use a series of benchmarks covering different tasks and video durations. NExT-QA~\cite{nextqa}, ActivityNetQA~\cite{activitynetqa}, and PerceptionTest~\cite{perceptiontest} are adopted to evaluate the question answering based on actions, object attributes, and object interactions. To assess more complex tasks that require reasoning and information extraction, we evaluate the model on benchmarks tailored for MLLMs, including VideoMME~\cite{videomme}, MLVU~\cite{mlvu}, LongVideoBench~\cite{longvideobench}, MVBench~\cite{videochat2mvbench}, and TempCompass~\cite{tempcompass}. We also test the model's ability for ego-centric videos on EgoSchema~\cite{egoschema}.

\begin{table}
\caption{\textbf{Statistics of the second-stage instruction tuning data.} }
\vspace{3pt}
\small
\addtolength{\tabcolsep}{-3pt}
\resizebox{\columnwidth}{!}{
\begin{tabular}{clc}
\toprule
Data Type                                 & Data Sources     & Number \\ \midrule
\multirow{7}{*}{Video} & LLaVA-Video-178k~\cite{zhang2024videoinstructiontuningsynthetic} & 1,395k \\
                                          & TGIF-QA~\cite{TGIF}          & 58k    \\
                                          & SthSthV2~\cite{sthsth}         & 40k    \\
                                          & Kinetics-710~\cite{kinetics710}     & 40k    \\
                                          & CLEVR~\cite{clevr}            & 20k    \\
                                          & VideoChat2~\cite{videochat2mvbench}       & 10k    \\
                                          & Ego4D~\cite{ego4d}            & 8k     \\ \midrule
\multirow{2}{*}{Image \& Pure Text}       & LLaVA-1.5~\cite{llava15}        & 665k   \\
                                          & LLaVA-OneVision~\cite{li2024llavaonevision} & 1,231k \\ \midrule
Sum                                       &                  & 3,467k \\ \bottomrule
\end{tabular}
}
\vspace{-0.5cm}
\label{tab:data-mixture-source}
\end{table}

\begin{table*}
\caption{\textbf{Comparisons with other video understanding MLLM.}  ``\#Tokens'' refers to the number of visual tokens fed into the LLM. *Indicates the maximum number of visual tokens. $\dagger$ Cross-attention-based MLLMs do not input visual tokens into the LLM.}
\vskip 0.1in
\centering
\addtolength{\tabcolsep}{-2pt}
\resizebox{2\columnwidth}{!}{
\begin{tabular}{l|lcc|cccccccccc}
\toprule
Method& LLM       &  \#Frames& \#Tokens&\rotatebox{60}{\textbf{\scriptsize{VideoMME}}} & \rotatebox{60}{\textbf{\scriptsize{VideoMME$_{\mathrm{sub}}$}}}& \rotatebox{60}{\textbf{\scriptsize{MLVU}}} & \rotatebox{60}{\textbf{\scriptsize{MVBench}}} & \rotatebox{60}{\textbf{\scriptsize{EgoSchema}}} & \rotatebox{60}{\textbf{\scriptsize{LongVideoBench}}} & \rotatebox{60}{\textbf{\scriptsize{Perception Test}}}& \rotatebox{60}{\textbf{\scriptsize{NExT-QA}}} & \rotatebox{60}{\textbf{\scriptsize{ActivityNetQA}}} & \rotatebox{60}{\textbf{\scriptsize{TempCompass}}}
\\ \midrule
\rowcolor[HTML]{EFEFEF} 
GPT-4o      & -&  -& -&59.9& 63.3& -& -& -& -& -& -& 57.0& 70.9
\\
\midrule
VILA~\cite{vila}        & Yi-34B& - & - & 60.1& 61.1& 56.7& -& 58.0& -& 54.0& 67.9& \underline{58.0} & -
\\
PLLaVA~\cite{pllava}      & Yi-34B& 16 & 2,304 &-& -& -& 58.1& -& 53.2& -& -& \textbf{60.9}& -
\\
LongVA~\cite{longva}      & Qwen2-7B & 128 & 18,432 &52.6& 54.3& 56.3& -& -& -& -& 68.3& 50.0& -
\\
IXC-2.5~\cite{internlmxcomposer25}     & InternLM2-7B & 32 & 12,800 &55.8& 58.8& 37.3& \textbf{69.1}& -& -& 34.4& 71.0& 52.8& -
\\
LLaVA-OV~\cite{li2024llavaonevision}    & Qwen2-7B& 64 & 12,545 &58.2& 61.5& 64.7& 56.7& \underline{60.1} & 56.5& 57.1& 79.4& 56.6& 64.8
\\
VideoLLaMA2~\cite{videollama2} & Qwen2-7B& 16 & 1,152 &47.9& 50.3& 32.7& 54.6& 51.7& -& 51.4& -& 50.2& -
\\
Kangoroo~\cite{kangaroo}    & LLaMA3-8B& 64 & 16,384 & 56.0& 57.6& 61.0& 61.1& \textbf{62.7}& 54.8& -& -& -& 61.3
\\
Oryx-MLLM~\cite{liu2024oryx}   & Qwen2-7B& 64 & 16,384* & 58.3& 62.6& \underline{67.5}& 63.9& -& 55.3& 68.6& 81.9& -& -
\\
mPLUG-Owl3~\cite{mplugowl3}  & Qwen2-7B& 8 & 0$^\dagger$ &53.5& -& -& 54.5& -& 52.1& -& 78.6& -& -
\\ \midrule
\multirow{2}{*}{\textbf{Slow-fast MLLM}}    & Qwen2-7B&  64& 1,296 &\underline{60.2}& \underline{63.0} & 67.3& \underline{68.9} & 59.2& \underline{56.6} & \textbf{70.3}& \textbf{83.5} & 54.8&          \textbf{68.9}\\
   & Qwen2-7B &  96     & 1,296 &\textbf{60.3}&       \textbf{63.4}&         \textbf{68.1}&      68.6&         59.8&           \textbf{58.0}&              \underline{70.2} &           \underline{83.1}&         54.5&          \underline{67.7}\\ 
   \bottomrule
\end{tabular}
}
\label{tab:main-comparison}
\end{table*}

\subsection{Comparison with Other Methods}

We compare our model with other state-of-the-art video understanding models in Table~\ref{tab:main-comparison}. For the proposed slow-fast MLLM, we use two settings which use $64$ and $96$ input frames. The temporal pooling strides are set to be $4$ and $6$ to generate \textit{fast tokens}.
Despite feeding limited visual tokens to the LLM, the 64-frame model achieves comparable or second-best performance across most benchmarks, while outperforming other models on VideoMME, LongVideoBench, PerceptionTest, NExT-QA, and TempCompass. Increasing the number of slow visual inputs to 96 frames further improves performance on most of the compared benchmarks, demonstrating the scalability and effectiveness of our approach.

It is worth noting that, compared to other competitive methods in Table~\ref{tab:main-comparison}, our model still has untapped potential. For instance, it could benefit from leveraging larger datasets and additional training stages, as seen in LLaVA-OneVision~\cite{li2024llavaonevision}, or incorporating more advanced vision encoders as Oryx-MLLM~\cite{liu2024oryx}.

\subsection{Ablation Study}
\label{sec:exp-ablation}
In this section, we conduct a series of ablation studies to analyze the architectural design choices. To better investigate the slow-fast mechanism, unless otherwise specified, we use direct strided sampling to obtain the fast visual tokens, ensuring that the text embedding relies on information from the slow visual tokens and ruling out the benefits of temporal pooling operations.

\begin{table}
\caption{\textbf{Comparisons between different decoder designs.} ``\#Frames'' denotes the input video frames to the LLM, ``64$\rightarrow$16'' denotes that $64$ frame is compressed into $16$ frames via temporal average pooling, and ``64/16'' denotes using $64$ slow frames and $16$ fast frames.  ``VMME'', ``MVB'', ``LongVid.'', and ``Ego.'' denote the VideoMME~\cite{videomme}, MVBench~\cite{videochat2mvbench}, LongVideoBench~\cite{longvideobench}, and EgoSchema~\cite{egoschema}, hereafter.}
\vskip 0.1in
\small
\addtolength{\tabcolsep}{-1pt}
\resizebox{\columnwidth}{!}{
\begin{tabular}{cc|cccccl}
\toprule
Architecture& \#Frames& VMME & MLVU & MVB & LongVid. & Ego. & Avg. \\ \midrule
Self-attn& 16 & 57.2& 58.5& 54.9& 48.9& 50.6&54.0
\\
Self-attn& 64$\rightarrow$16 & 58.9& 64.5& 55.7& 51.2& 50.3&56.1
\\
Cross-attn& 64 & 50.8& 53.2& 49.6& 47.7& 46.2&49.5
\\
Slow-Fast & 64/16 & 58.5& 62.5& 59.5& 53.5& 57.5&58.3
\\ 
Slow-Fast & 64/64$\rightarrow$16 & \textbf{60.3}& \textbf{65.9}& \textbf{60.6}& \textbf{55.4}& \textbf{61.0}&\textbf{60.7}
\\ \bottomrule
\end{tabular}
}
\vspace{-0.3cm}
\label{tab:comparison-different-baselines}
\end{table}

\noindent \textbf{Different architectures.}
First, we compare the proposed slow-fast architecture with the self-attention and cross-attention architecture. For self-attention models, we use two settings: 1) using 16 frames as input and 2) compressing 64 frames into 16 frames with temporal pooling. For the cross-attention model, we use 64 frames as input.
We simply remove the fast visual tokens from the proposed slow-fast architecture without other modifications to construct the cross-attention baseline. 
For slow-fast MLLM, we use $64$ input frames and adopt two different ways to generate the fast visual tokens: uniformly sample 16 frames from the slow visual tokens and compress the slow-visual tokens into 16 frames with temporal pooling. 
As shown in Table~\ref{tab:comparison-different-baselines}, we can conclude the following: 1) The slow-fast architecture demonstrates clear advantages over both the self-attention and cross-attention baselines; 2) Comparing row 1 with row 4, adding slow visual tokens significantly improves performance on EgoSchema (13.6\%), MVBench (8.2\%), and LongVideoBench (8.9\%); 
3) Comparing row 3 with row 4, removing the fast tokens causes a very significant performance drop, demonstrating the necessity of adding the context directly into the LLM;
4) Comparing row 2 with row 4, slow-fast architecture achieves better average scores than direct frame compression.
5) When generating the fast visual tokens with temporal pooling instead of direct sampling, row 5 achieves the best performance across all benchmarks, indicating that slow-fast architecture can further benefit from more advanced token compression.

\noindent \textbf{Cross-attention integration.} 
As discussed in Sec.\ref{sec:hybrid-decoder-layer}, cross-attention can either be integrated into a stand-alone decoder layer, similar to Flamingo~\cite{flamingo}, or inserted into the decoder layer in parallel with the self-attention, denoted as ``Hybrid''. As shown in Table~\ref{tab:comparison-different-decoder-design}, the hybrid implementation achieves the best overall performance.

\begin{table}
\caption{\textbf{Comparisons between different decoder design.} ``Stand-alone'' denotes that the cross-attention layer is integrated into a separate decoder layer while hybrid denotes that the cross-attention layer is inserted between the self-attention and the MLP in the standard decoder.}
\vskip 0.15in
\small
\addtolength{\tabcolsep}{-1pt}
\resizebox{\columnwidth}{!}{
\begin{tabular}{cc|cccccl}
\toprule
Decoder     & FFN   & VMME& MLVU& MVB& LongVid.& Ego.& Avg.\\ \midrule
Standard    & -     & 57.2& 58.5& 54.9& 48.9& 50.6&54.0
\\
Stand-alone & \ding{51}  & 57.3& 60.6& 57.9& 51.7& 56.4&56.8
\\
Stand-alone & \ding{53} & 56.9& 59.2& 58.0& 51.6& 55.9&56.3
\\
Hybrid      & -     & \textbf{58.5}& \textbf{62.5}& \textbf{59.5}& \textbf{53.5}& \textbf{57.5}&\textbf{58.3}
\\ \bottomrule
\end{tabular}
}
\vspace{-0.3cm}
\label{tab:comparison-different-decoder-design}
\end{table}

\noindent \textbf{Gate.} 
We compare three different gates: 1) \textbf{Static} gate using a learnable scalar for all text tokens; 2) \textbf{Dyn.}, dynamic gate used our method which predicts a gate value for each text token based on its embedding; 3) \textbf{C-Dyn.}, channel-wise dynamic gate further predicts separate values for each channel of the text token, as in mPLUG-Owl3~\cite{mplugowl3}. 
Both \textbf{Dyn.} and \textbf{C-Dyn.} gate use the learnable warm-up factor discussed in Sec.~\ref{sec:hybrid-decoder-layer} to prevent loss spikes during the pre-training.
As shown in Table~\ref{tab:comparison-different-gates-init}, comparing rows 1 and 4, the token-wise dynamic gate achieves consistently better performance across benchmarks compared to the static gate, with a particularly notable improvement of $9\%$ on VideoMME with subtitles. This highlights the ability of the dynamic gate to reduce distraction from unrelated visual information into particular text embeddings, especially for lengthy textual inputs. However, the channel-wise dynamic gate does not provide additional benefits, suggesting that per-token modulation is sufficient for optimizing the model's performance.

\noindent \textbf{Key-Value projection initialization.} 
We compare three initialization strategies for the cross-attention key-value projection layer: 1) \textbf{Rand.}: random initialization; 2) \textbf{Share}: sharing key-value projections with the self-attention layer; 3) \textbf{Copy}: initializing projection weights from the corresponding self-attention layer.
Table~\ref{tab:comparison-different-gates-init} shows both ``share'' and ``copy'' initialization is significantly better than random initialization, while ``copy'' gives a more pronounced $4\%$ improvement on the average performance, showing that leveraging pre-trained weights is important.

\begin{table}
\caption{\textbf{Comparisons of different gate designs and weight initializations.}}
\vskip 0.1in
\addtolength{\tabcolsep}{-3pt}
\resizebox{\columnwidth}{!}{
\begin{tabular}{cc|ccccccc}
\toprule
Gate   & Init. & VMME& VMME$_{\mathrm{sub}}$& MLVU& MVB & LongVid.& Ego. & Avg.\\ \midrule
Static & Copy  & 57.0& 57.4& 59.6& 59.1& 52.5& 53.9& 56.4
 \\
Dyn.   & Rand. & 56.0& 58.3& 59.0& 57.8& 51.9& 54.3& 55.8
 \\
Dyn.   & Share & 58.1& 61.6& 60.7& \textbf{60.3}& 51.8& 56.1& 57.4
 \\
Dyn.   & Copy  & \textbf{58.5}& \textbf{62.6}& \textbf{62.5}& 59.5& \textbf{53.5}& \textbf{57.5}& \textbf{58.3}
 \\
C-Dyn. & Copy  & 57.6& 61.6& 57.2& 60.1& 50.9& 54.6& 56.1
 \\ \bottomrule
\end{tabular}
}
\vspace{-0.3cm}
\label{tab:comparison-different-gates-init}
\end{table}

\noindent \textbf{Query for cross-attention.} 
Considering the efficiency, only the text tokens in the input sequence are allowed to attend to the slow visual tokens in our implementation. Here we compare two different settings: 1) \textbf{All}, all tokens are used as the query or 2) \textbf{Visual}, only the fast visual tokens are used as queries. As shown in Table~\ref{tab:comparison-different-query}, limiting attention to text tokens yields the best performance while the all-token-as-query strategy slightly falls.
Allowing only visual tokens to query the slow visual tokens results in the lowest average score, showing that sampling visual information conditioned on text instructions is important.

\begin{table}
\caption{\textbf{Comparisons between different query settings.} ``All'' means that both the fast visual tokens and text tokens can attend to the slow visual tokens.}
\vskip 0.1in
\centering
\small
\addtolength{\tabcolsep}{2pt}
\resizebox{\columnwidth}{!}{
\begin{tabular}{c|cccccc}
\toprule
Query  & VMME& MLVU& MVB& LongVid.& Ego. & Avg. \\ \midrule
All    & \textbf{58.6}& 60.2& 58.8& 54.9& 53.0& 57.1
\\
Visual & 57.3& 59.2& \textbf{59.5}& 52.4& \textbf{54.5}& 56.6
\\
Text   & 58.5& \textbf{62.5}& \textbf{59.5}& \textbf{57.5}& 53.5& \textbf{58.3}
\\ \bottomrule
\end{tabular}
}
\vspace{-.3cm}
\label{tab:comparison-different-query}
\end{table}

\noindent \textbf{The impact of frame number.}
Here we compare different input frame counts and compression ratios between the fast and slow visual tokens. If the fast visual tokens are generated by compressing $m$ frames using stride-$n$ sampling (temporal pooling), this configuration is denoted as ``$m$-s$n$'' (or ``$m$-p$n$'' for pooling).
As shown in Table~\ref{tab:comparison-different-input-frames}, increasing the number of input frames consistently improves the performance, especially for benchmarks requiring long video comprehension. For example, according to rows 3 and 4, increasing input frames from 64 to 96 improves performance on the \texttt{long} subset of VideoMME ($+2\%$), EgoSchema ($+14\%$), and on average ($+3\%$).
The number of fast visual tokens is $1,296$ for both settings, which shows the slow-fast architecture's ability to process long video sequences without increasing the input sequence length for the LLM.

\begin{table}
\caption{\textbf{Impact of different input frame numbers.} If the fast visual tokens are generated by compressing $m$ frames using stride-$n$ sampling (temporal pooling), this configuration is denoted as ``$m$-s$n$'' (or ``$m$-p$n$'' for pooling).}
\vskip 0.1in
\addtolength{\tabcolsep}{-2pt}
\resizebox{\columnwidth}{!}{
\begin{tabular}{c|ccccccc}
\toprule
Config & VMME& VMME$_{\mathrm{L}}$ & MLVU & MVB & LongVid. & Ego. & Avg.\\ \midrule
48-s3  & 57.2& 49.1& 59.6& 58.6& 51.9& 53.1& 56.1
\\
64-s4  & 58.5& 50.4& 62.5& 59.5& 53.5& 57.5& 58.3
\\
64-p4  & 60.3& 50.2& 65.9& 60.3& 55.1& 57.4& 59.8
\\
96-p6  & 61.1& 51.3& 66.5& 60.2& 55.3& 65.5& 61.7
\\
128-s2p4 & 61.5& 52.2& 66.7& 60.6& 57.3& 67.0& 62.6\\ \bottomrule
\end{tabular}
}
\label{tab:comparison-different-input-frames}
\end{table}

\subsection{Efficiency Analysis}
Table~\ref{tab:comparison-efficiency} reports the floating-point operations (FLOPs) of the first forward pass of different architectures. Computation on the vision encoder is not counted. As the input frame grows, slow-fast architecture only introduces marginal additional computation compared to the self-attention baseline. For example, when increasing the input frames from 16 to 64 and 96, the slow-fast architecture adds only 0.16 and 0.24 TFLOPs, accounting for just 0.8\% and 1.2\% of the total LLM computation. In contrast, the self-attention baseline has more than a fourfold (64 frame) and sixfold (96 frame) increase in computation. Furthermore, as each hybrid layer only has two lightweight projection layers and a dynamic gate, the total parameter count increases by just 0.2\%, highlighting the parameter efficiency of the proposed design.

\begin{figure*}[!t]
   \centering
   \includegraphics[width=1.0\linewidth]{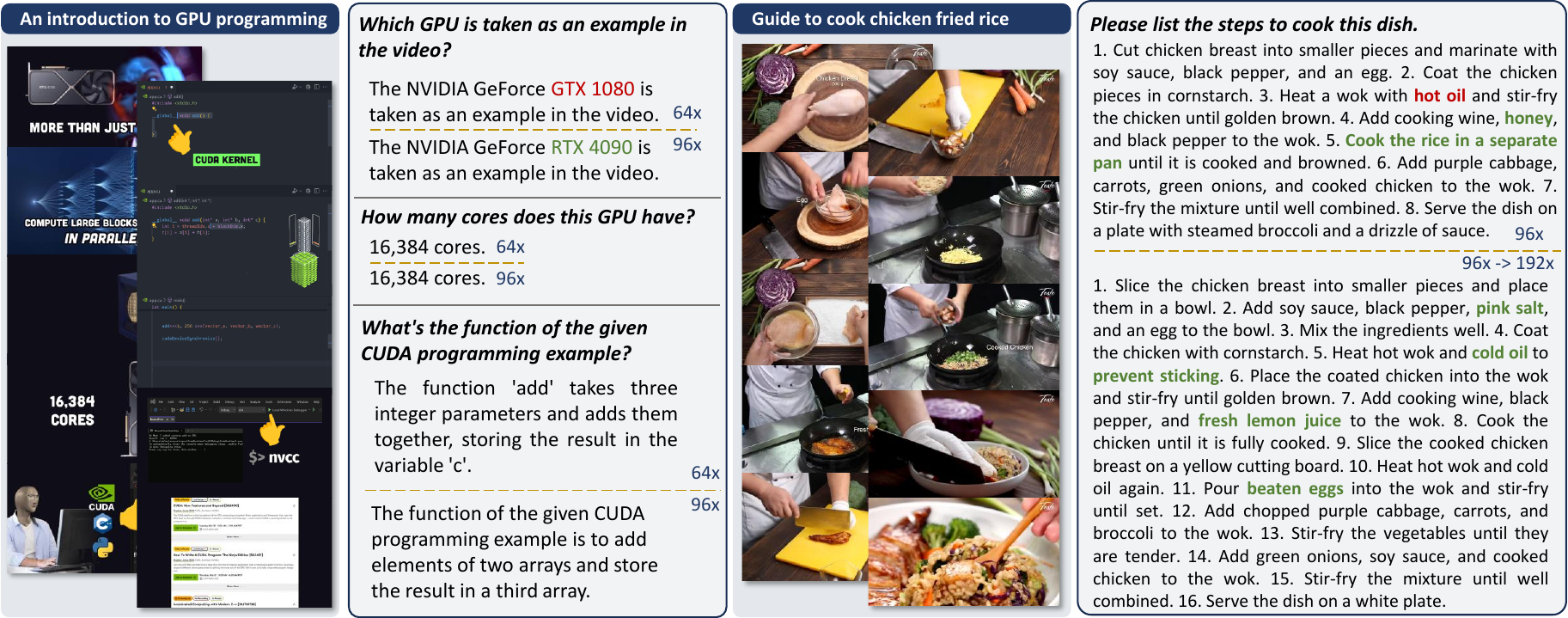}
   \vspace{-0.4cm}
   \caption{\textbf{Qualitative examples and comparisons between different input frame numbers.}
   For the video on the left, models trained and tested with 64 and 96 frames are compared, denoted as ``64x'' and ``96x''. In the video on the right, we further apply test time augmentation by increasing the input frames to 192. \textit{More comparisons are available in the supplement.}
   }
   \label{fig:qualitative-examples}
\vspace{-.3cm}
\end{figure*}

\begin{table}
\caption{\textbf{Efficiency comparisons between different architectures.} ``\# Tokens'' denotes the number of input visual tokens for the LLMs. ``64/16'' denotes that the numbers of slow and fast frames are 64 and 16.}
\vskip 0.1in
\addtolength{\tabcolsep}{-2pt}
\resizebox{\columnwidth}{!}{
\begin{tabular}{c|ccccc}
\toprule
Arch. & \# Frames & \# Tokens & \# Params & LLM TFLOPs & CA TFLOPs \\ \midrule
Self-attn    & 16       & 1296     & 8.48B     & 19.64
& -         \\
Self-attn    & 32       & 2592     & 8.48B     & 40.21
& -         \\
Self-attn    & 64       & 5184     & 8.48B     & 85.57
& -         \\
Self-attn    & 96       & 7776     & 8.48B     & 136.16
& -         \\
Slow-Fast       & 64/16       & 1296     & 8.50B     & 19.80
& 0.16      \\
Slow-Fast       & 96/16       & 1296     & 8.50B     & 19.88
& 0.24      \\ \bottomrule
\end{tabular}
}
\vspace{-.4cm}
\label{tab:comparison-efficiency}
\end{table}

\subsection{Qualitative Results}
\noindent \textbf{Open-ended conversations.} 
Fig.~\ref{fig:qualitative-examples} illustrates real-world scenarios that require precise information extraction, OCR, reasoning, and summarization abilities. In the GPU programming example on the left, the model recognizes the text within the video to answer questions accurately. It can also read the program step by step and summarize the function by integrating information across the video content. In the cooking video on the right side, the model lists the required ingredients and operations, showcasing its structured understanding and summarization ability.

Fig.~\ref{fig:qualitative-examples} also shows the benefits of more input frames intuitively.
For example, with only 64 frames, some critical details are lost due to frame sampling, leading to errors or hallucinations. In the second example, we increase the input frame count from 96 to 192 by modifying the sampling stride from 1 to 2, ensuring that the fast frame count remains fixed at $16$. Meanwhile, the temporal pooling stride is kept at 6. In this way, more details are captured, such as “\textit{pink salt}” and “\textit{beaten eggs}”, and corrected errors, \textit{e.g.}, revising “\textit{hot oil}” to “\textit{cold oil to prevent sticking}”. However, direct interpolation also led to omissions of certain details toward the end of the video. A plausible reason is that the length of the slow visual tokens doubles compared to the training.

\begin{figure}[!t]
   \centering
   \includegraphics[width=1.0\linewidth]{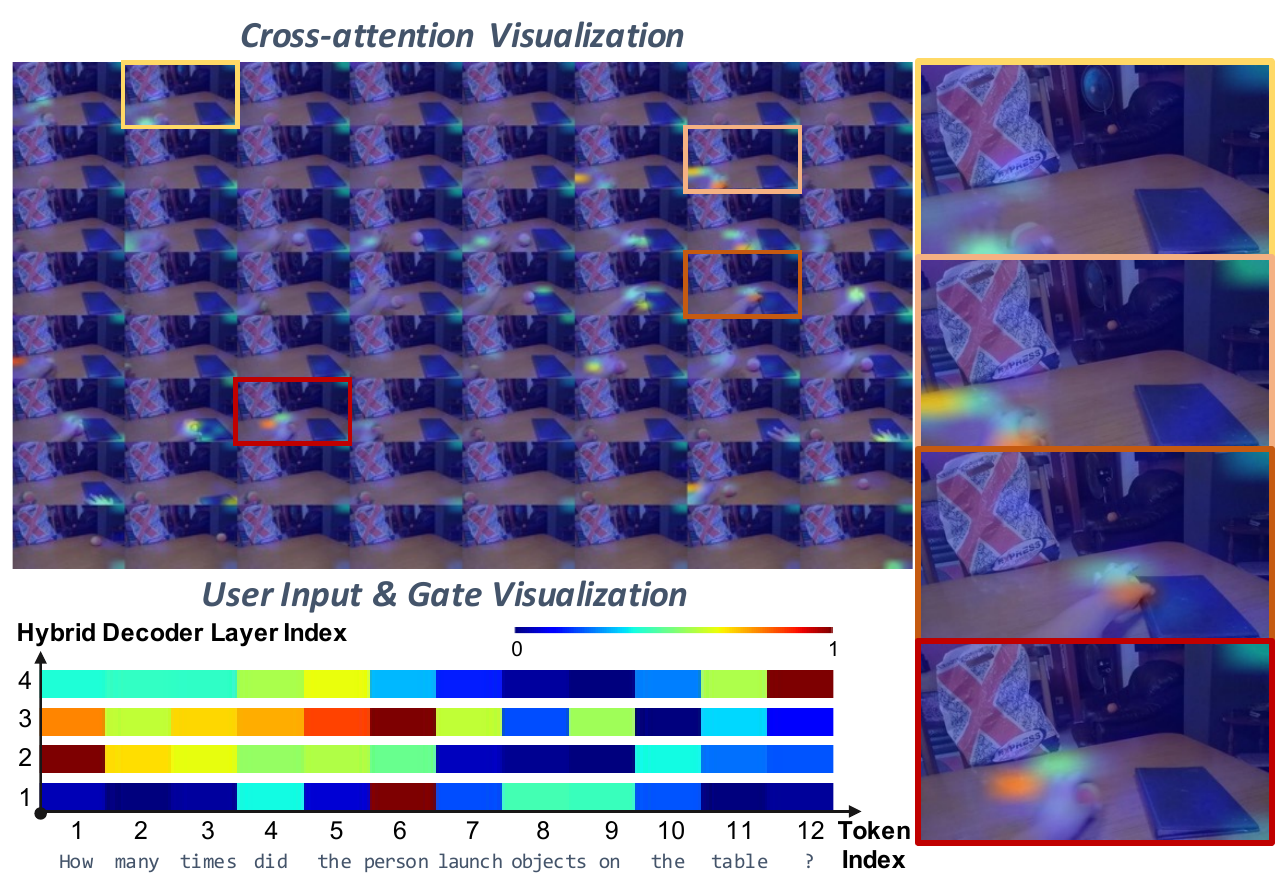}
   \vspace{-0.5cm}
   \caption{\textbf{Visualizations of the cross-attention map and the dynamic gate in the hybrid decoder.} The cross-attention maps are averaged across different decoder layers, text tokens, and attention heads. The absolute value of the dynamic gate from all the four hybrid decoder layers are visualized.
   }
   \label{fig:attn-gate-visualization}  
\vspace{-.3cm}
\end{figure}

\noindent \textbf{Cross-attention and dynamic gate.} 
We visualize the cross-attention maps in Fig.~\ref{fig:attn-gate-visualization}, which are averaged across different attention heads, tokens, and decoder layers. When the instruction is “How many times did the person launch the objects on the table?”, the attention focuses on the hands and objects in the video, particularly during moments of significant motion of the hands and objects. This example shows how the text embeddings dynamically retrieve relevant visual clues. Additionally, the gate value for each text token is visualized in Fig.~\ref{fig:attn-gate-visualization}, revealing that different layers prioritize distinct tokens.

\section{Conclusion}

In this paper, we propose a slow-fast architecture for video understanding inspired by the human video question-answering process. 
The highly compressed video representation, referred to as the fast visual tokens are fed into the LLM serving as the preview context, while the text embeddings can interact with uncompressed video representation, enabling text-aware visual information extraction with linear complexity.
Experiments demonstrate that our models significantly outperform conventional self-attention-only architecture both in performance and the efficiency to process long video inputs.
We hope this work inspires further innovations in the architectural design of multi-modal large language models, enabling more dynamic and efficient interactions between language models and other modalities.

\section*{Acknowledgment}
The authors would like to thank NVIDIA and Georgia Tech for their support. This work was in part supported by the National Science Foundation (NSF) and the Institute of Education Sciences under the National AI Institute for Exceptional Education Award (\#2229873) and the NSF CAREER Award (\#2239840).



\bibliography{reference}
\bibliographystyle{icml2025}

\newpage
\appendix
\onecolumn

This appendix contains the following contents:
\begin{enumerate}
    \item More qualitative results in Sec.~\ref{supp-sec:qualitative-results}.
    \item More details on the model architecture in Sec.~\ref{supp-sec:archiecture-details}.
    \item Evaluation prompt for all the tested benchmarks in Sec.~\ref{supp-sec:evaluation-prompt}.
\end{enumerate}

\section{More Qualitative Results}
\label{supp-sec:qualitative-results}
\begin{figure*}[htbp]
   \centering
   \includegraphics[width=1.0\linewidth]{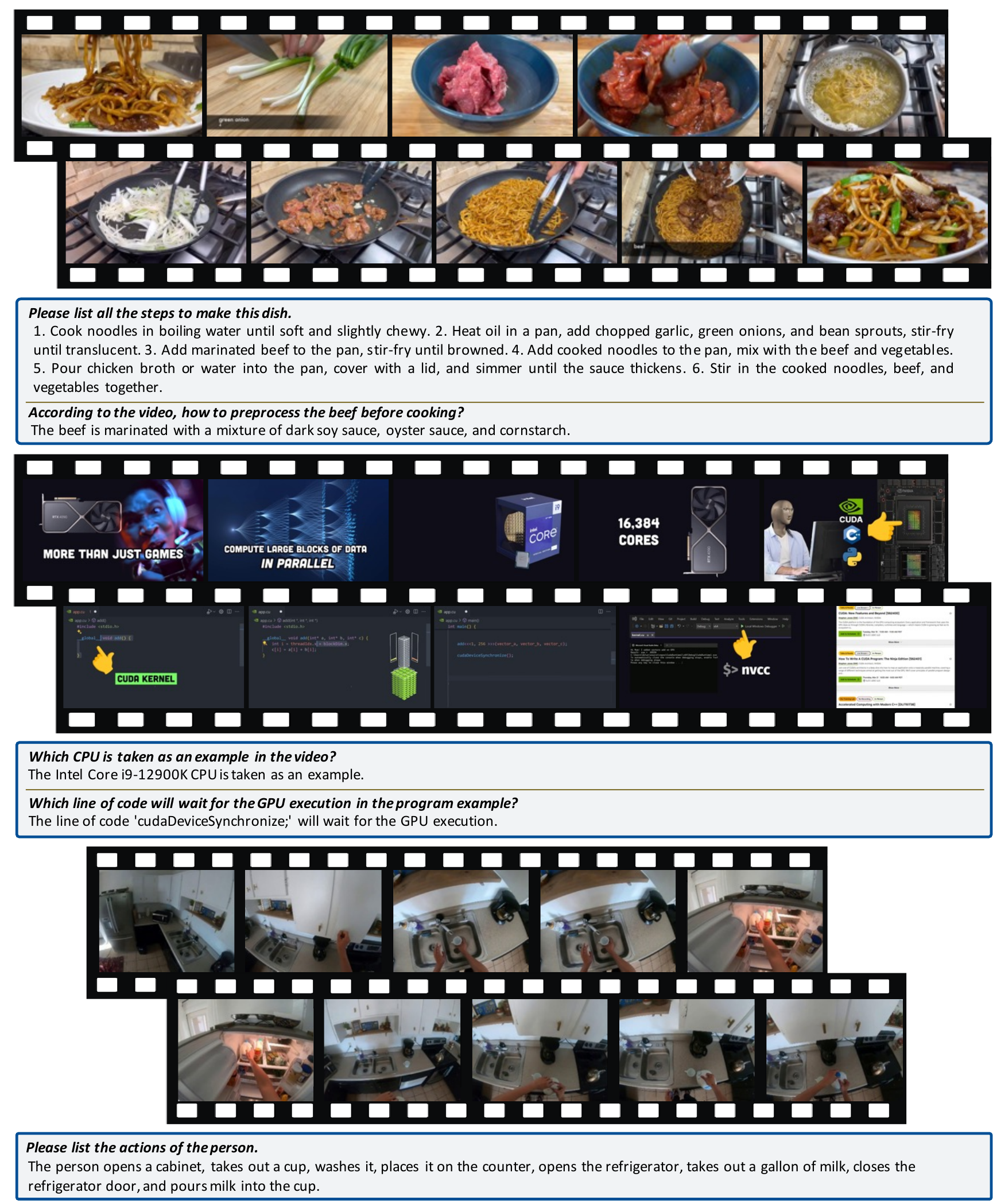}
   \vspace{-0.6cm}
   \caption{\textbf{More qualitative examples of our model.}}
   \label{supp-fig:more-qualitative-examples}
\vspace{-.3cm}
\end{figure*}

More qualitative results of our model are presented in Fig.~\ref{supp-fig:more-qualitative-examples}, including question answering and summarization tasks.

\begin{figure*}[htbp]
   \centering
   \includegraphics[width=1.0\linewidth]{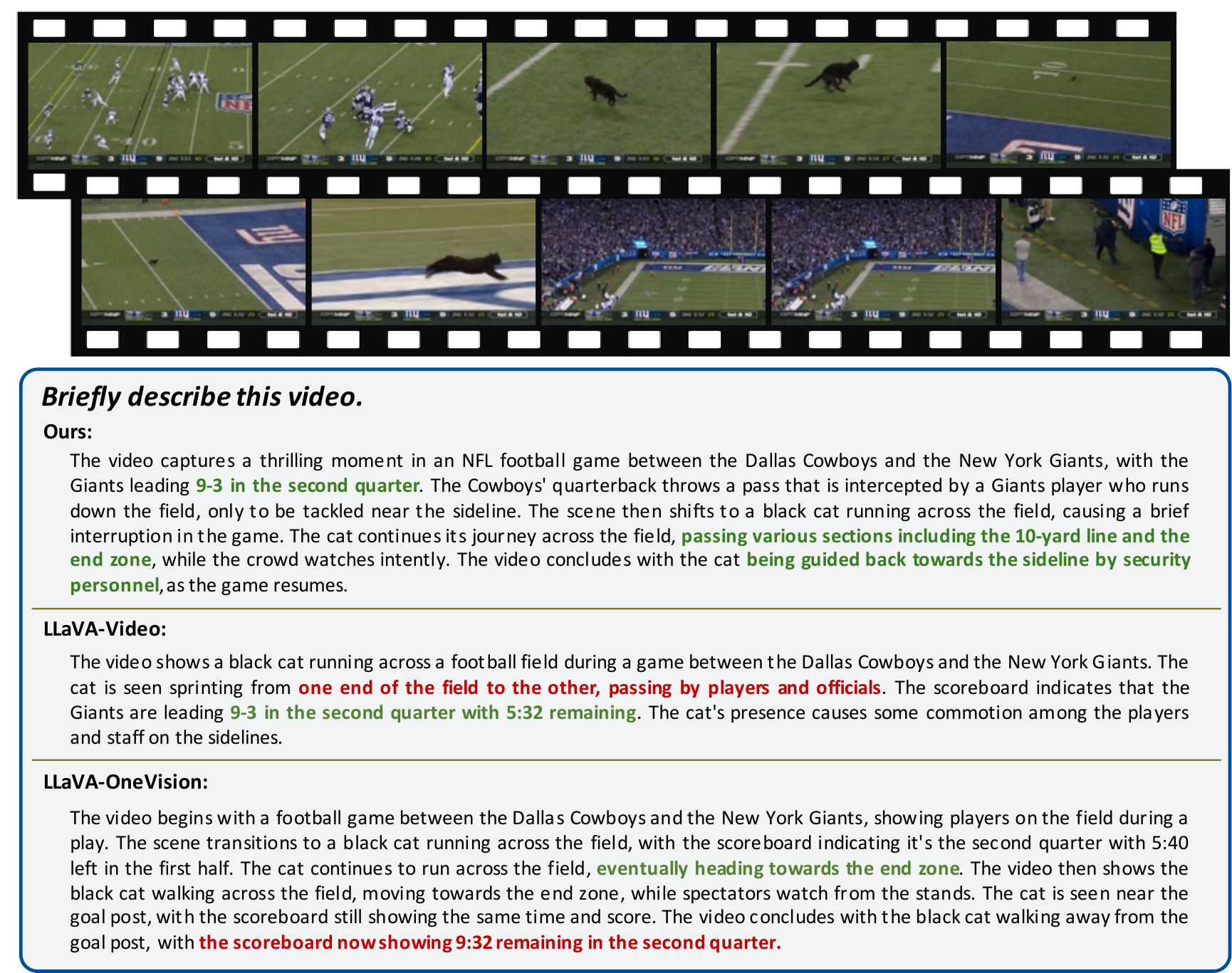}
   \vspace{-0.6cm}
   \caption{\textbf{Comparison with other state-of-the-art Video MLLMs on video description.}}
   \label{supp-fig:qualitative-comparison-description}
\vspace{-.3cm}
\end{figure*}

\begin{figure*}[htbp]
   \centering
   \includegraphics[width=1.0\linewidth]{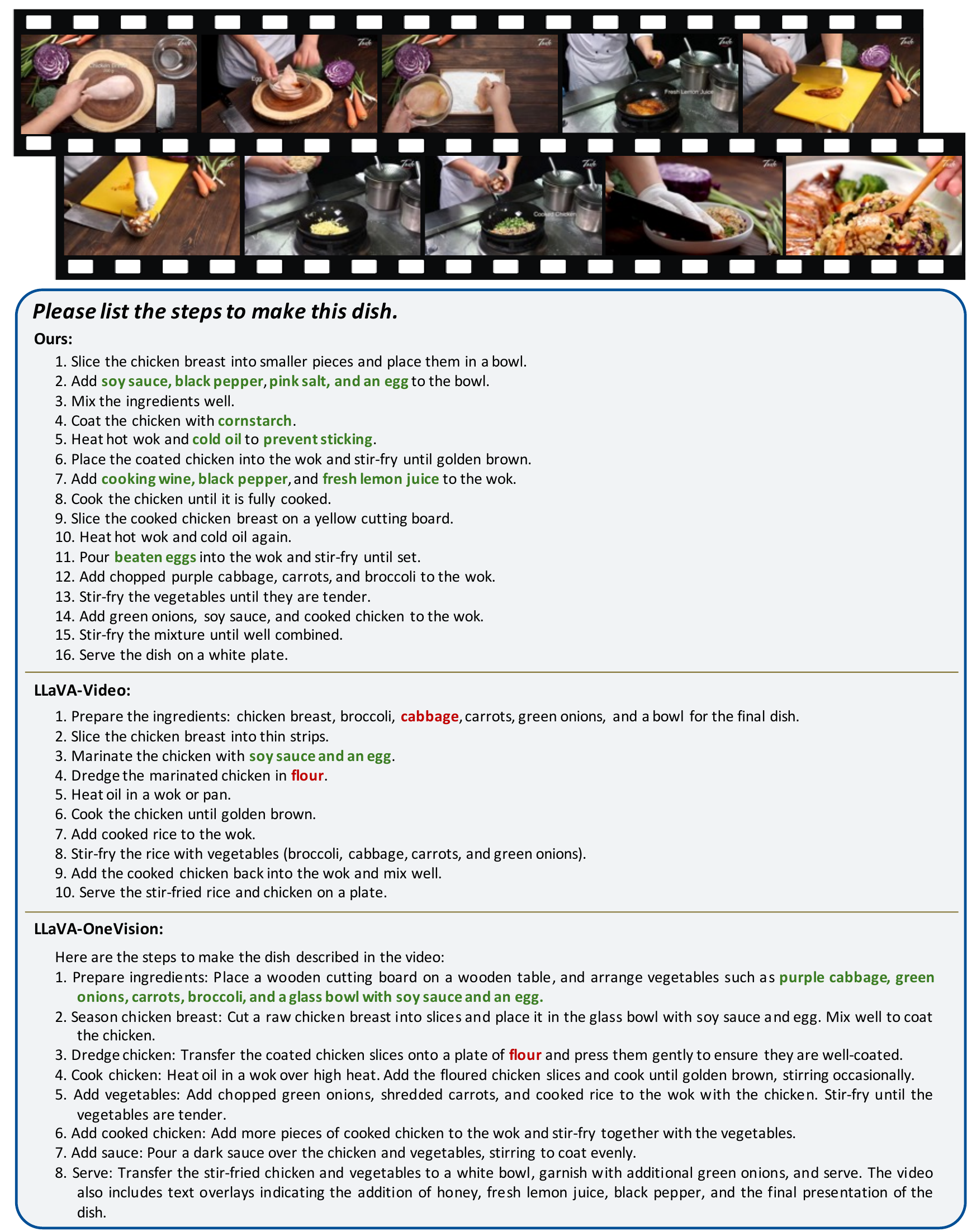}
   \vspace{-0.6cm}
   \caption{\textbf{Comparison with other state-of-the-art Video MLLMs on information summarization.}}
   \label{supp-fig:qualitative-comparison-summarize}
\vspace{-.3cm}
\end{figure*}

\begin{figure*}[htbp]
   \centering
   \includegraphics[width=1.0\linewidth]{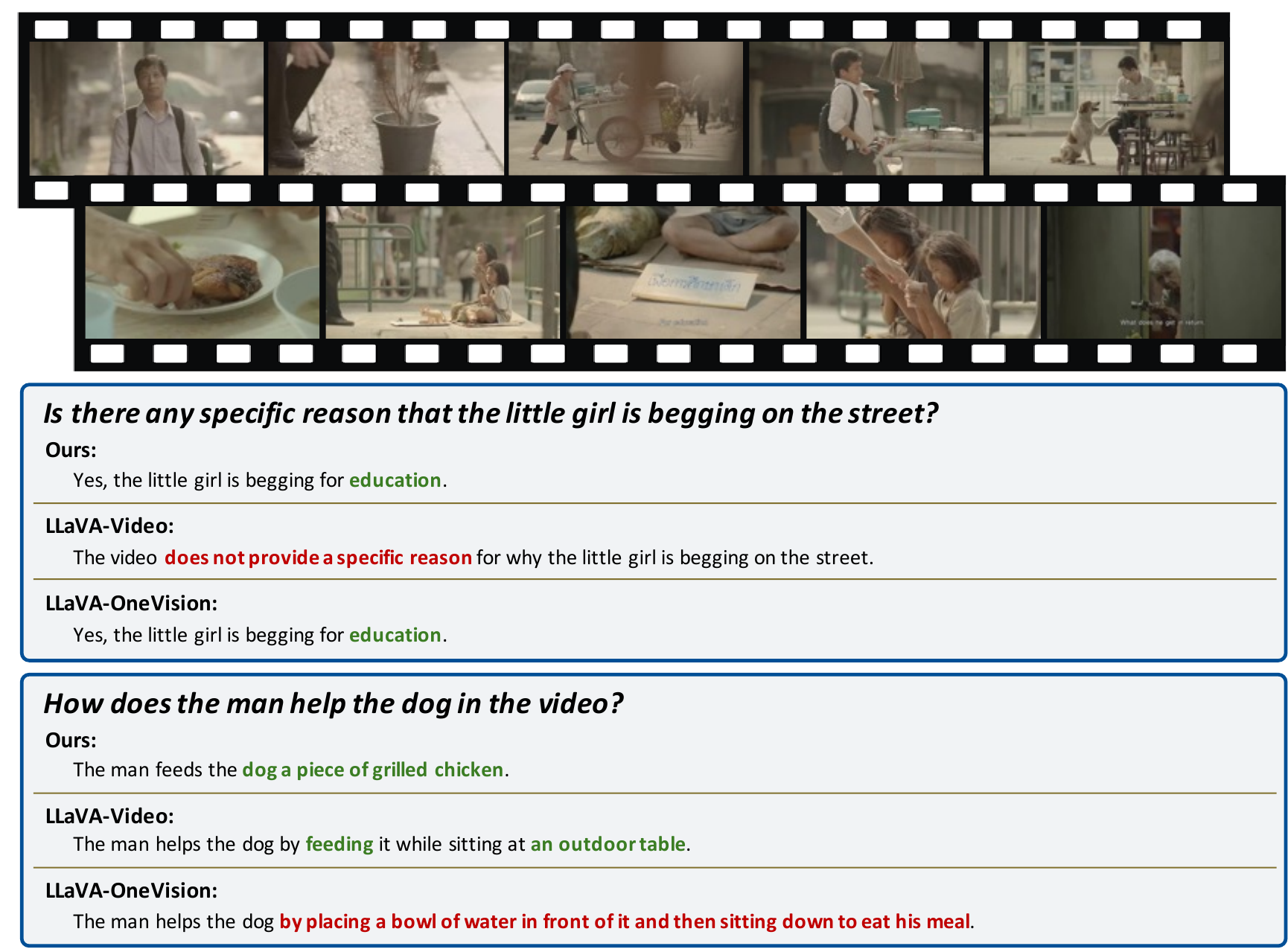}
   \vspace{-0.6cm}
   \caption{\textbf{Comparison with other state-of-the-art Video MLLMs on video question understanding.}}
   \label{supp-fig:qualitative-comparison-qa}
\vspace{-.3cm}
\end{figure*}

Additionally, we provide qualitative comparisons with state-of-the-art methods, LLaVA-OneVision~\cite{li2024llavaonevision} and LLaVA-Video~\cite{zhang2024videoinstructiontuningsynthetic}. Fig.~\ref{supp-fig:qualitative-comparison-description}, Fig.~\ref{supp-fig:qualitative-comparison-summarize}, and Fig.~\ref{supp-fig:qualitative-comparison-qa} illustrate examples for video description, summarization, and question answering, respectively. Our model generates more accurate details and maintains better alignment with the video content, whereas the compared methods exhibit hallucinations and omit certain details.

\newpage
    
\section{Model Architecture Details}
\label{supp-sec:archiecture-details}
Here, we provide more details of the architecture design, including the multi-modal projector, the fast visual token compression process, and the hybrid decoder layer.

\noindent \textbf{Multi-modal projector.} We adopt the multi-modal projector design from LLaVA 1.5~\cite{llava15}, which consists of two linear projection layers with a GELU activation in between.

\noindent \textbf{Fast visual token compression.} Below, we detail the process of obtaining fast visual tokens. The input of this process is: 1) feature of $n$ video frames with the shape $n \times c \times H \times W$; 2) temporal sampling stride $k$; 3) temporal pooling stride $t$; and 4) minimum number of fast frames $m$.

The fast visual tokens are obtained through the following steps:
\begin{enumerate}
    \item Zero padding $n$ video frames on temporal axis to $n'$ frames, ensuring the total length is divisible by $k\cdot t$.
    \item Uniformly sample $n'' = \mathrm{max}(\lfloor \frac{n'}{k}\rfloor, m)$ frames on the padded video feature.
    \item Apply adaptive average pooling along the temporal axis to further compress the sampled feature to $\mathrm{max}(\lfloor \frac{n''}{t}\rfloor, m)$ frames.
    \item Flatten the compressed features along the spatial dimensions to generate the fast visual tokens.
\end{enumerate}

\begin{figure*}[!h]
   \centering
   \includegraphics[width=0.6\linewidth]{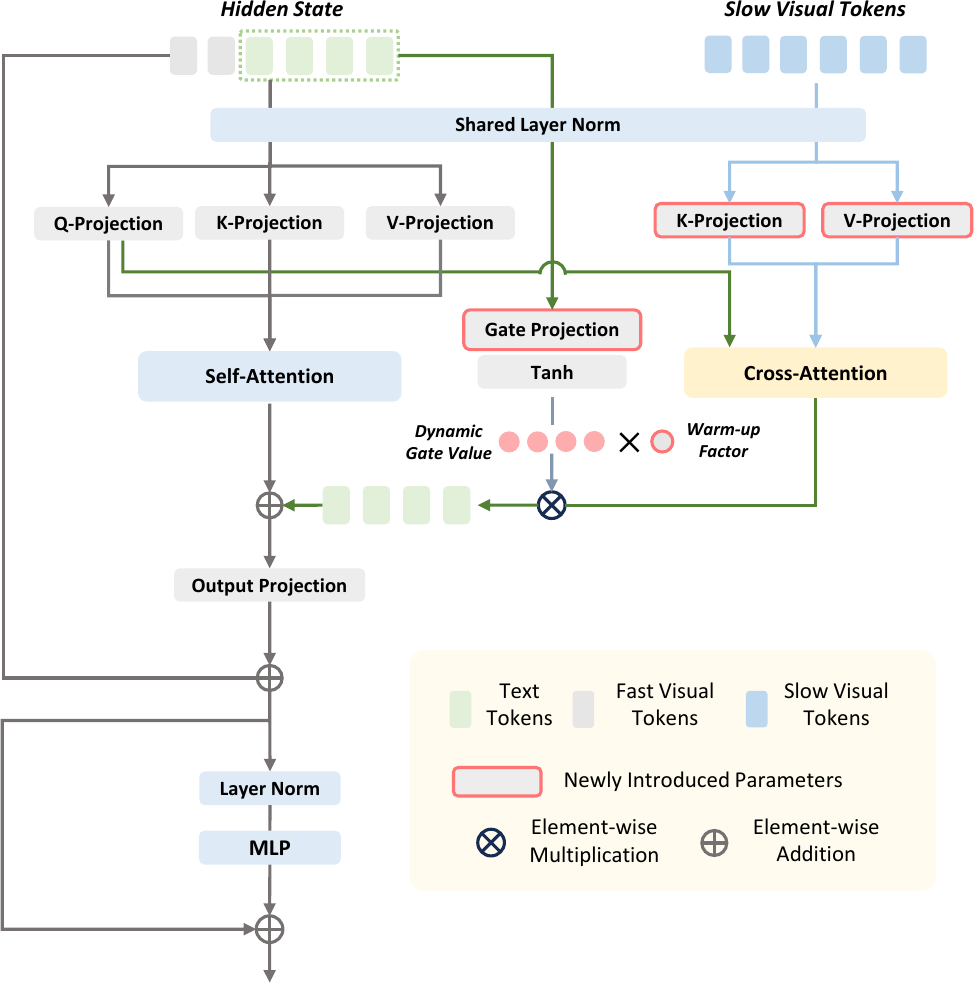}
   \vspace{-0.3cm}
   \caption{\textbf{Detailed architecture illustration of the hybrid decoder layer.}}
   \label{supp-fig:hybrid-decoder-layer}  
\end{figure*}

\noindent \textbf{Hybrid decoder layer.} The detailed architecture of the hybrid decoder layer is illustrated in Fig.~\ref{supp-fig:hybrid-decoder-layer}. Newly introduced parameters and modules are highlighted with pink outlines, while the remaining components originate from the pre-trained decoder layer.

\newpage

\section{Prompts for Evaluation}
\label{supp-sec:evaluation-prompt}
Here we specify the prompts we use for different benchmarks. For multi-choice selection benchmarks, including: VideoMME (w/w.o. subtitles), MVBench, MLVU, LongVideoBench, EgoSchema, PerceptionTest, and the multi-choice split of TempCompass, we unify the prompt for testing as the format below:
\begin{tcolorbox}
\textbf{Evaluation prompt for multi-choice question answering benchmarks.} \\
{\color{lightblue}{$<$Video$>$}} \\
Select the best answer to the following multiple-choice question based on the video. \\
{\color{lightblue}{$<$Question$>$}} \\
A. {\color{lightblue}$<$Option 1$>$} \\
B. {\color{lightblue}$<$Option 2$>$} \\
C. {\color{lightblue}$<$Option 3$>$} \\
D. {\color{lightblue}$<$Option 4$>$} \\
E. {\color{lightblue}$<$Option 5$>$} \\
\textcolor[rgb]{0.5,0.5,0.5}{Other options $\cdots$} \\
Answer with the option's letter from the given choices directly.
\end{tcolorbox}

Specifically, for VideoMME with subtitles, we use the following prompt to integrate the subtitle information.
\begin{tcolorbox}
\textbf{Evaluation prompt for VideoMME with subtitiles.} \\
{\color{lightblue}{$<$Video$>$}} \\
This video's subtitles are listed below: \\
{\color{lightblue}{$<$Subtitles$>$}} \\
Select the best answer to the following multiple-choice question based on the video and the subtitles. \\
{\color{lightblue}{$<$Question$>$}} \\
A. {\color{lightblue}$<$Option 1$>$} \\
B. {\color{lightblue}$<$Option 2$>$} \\
C. {\color{lightblue}$<$Option 3$>$} \\
D. {\color{lightblue}$<$Option 4$>$} \\
\textcolor[rgb]{0.5,0.5,0.5}{Other options $\cdots$} \\
Answer with the option's letter from the given choices directly.
\end{tcolorbox}

For the open-ended benchmark ActivityNet, we use the following format as below:

\begin{tcolorbox}
\textbf{Evaluation prompt for ActivityNet-QA.} \\
{\color{lightblue}{$<$Video$>$}} \\
{\color{lightblue}{$<$Question$>$}} \\
Answer the question using a single word or phrase.
\end{tcolorbox}


\end{document}